\documentclass[conference]{IEEEtran}
\IEEEoverridecommandlockouts
\usepackage{cite}
\usepackage{amsmath,amssymb,amsfonts}
\usepackage{algorithmic}
\usepackage{graphicx}
\usepackage{textcomp}
\usepackage{xcolor}
\usepackage{url}
\def\BibTeX{{\rm B\kern-.05em{\sc i\kern-.025em b}\kern-.08em
    T\kern-.1667em\lower.7ex\hbox{E}\kern-.125emX}}
\begin{document}

\title{Fourier-Based GAN Fingerprint Detection using ResNet50}

\author{\IEEEauthorblockN{Sai Teja Erukude}
\IEEEauthorblockA{\textit{Department of Computer Science} \\
\textit{Kansas State University}\\
Manhattan, USA \\
erukude.saiteja@gmail.com}
\and
\IEEEauthorblockN{Viswa Chaitanya Marella}
\IEEEauthorblockA{\textit{College of Business Administration} \\
\textit{Kansas State University}\\
Manhattan, USA \\
viswachaitanyamarella@gmail.com}
\and
\IEEEauthorblockN{Suhasnadh Reddy Veluru}
\IEEEauthorblockA{\textit{College of Business Administration} \\
\textit{Kansas State University}\\
Manhattan, USA \\
suhasnadhreddyveluru@gmail.com}
}

\maketitle

\begin{abstract}

The rapid rise of photorealistic images produced from Generative Adversarial Networks (GANs) poses a serious challenge for image forensics and industrial systems requiring reliable content authenticity. This paper uses frequency-domain analysis combined with deep learning to solve the problem of distinguishing StyleGAN-generated images from real ones. Specifically, a two-dimensional Discrete Fourier Transform (2D DFT) was applied to transform images into the Fourier domain, where subtle periodic artifacts become detectable. A ResNet50 neural network is trained on these transformed images to differentiate between real and synthetic ones. The experiments demonstrate that the frequency-domain model achieves a 92.8 percent and an AUC of 0.95, significantly outperforming the equivalent model trained on raw spatial-domain images. These results indicate that the GAN-generated images have unique frequency-domain signatures or ``fingerprints". The method proposed highlights the industrial potential of combining signal processing techniques and deep learning to enhance digital forensics and strengthen the trustworthiness of industrial AI systems. 

\end{abstract}


\section{Introduction}
Photorealistic fake images produced by GANs have become increasingly difficult for humans to distinguish from real photos. Intriguingly, state-of-the-art detection systems can still successfully differentiate GAN-generated images from originals. These detection systems are often trained to exploit subtle artifacts or ``fingerprints" that are embedded during the image acquisition phase of the GAN \cite{Neves2022}. Indeed, previous studies suggest that GANs exhibit unique fingerprinting behavior. For instance, GAN images contain unique noise features or consistent spectral signatures, while real photos tend to have noise features unique to the individual format and time of acquisition \cite{wang2023generalgangeneratedimagedetection}. Developing detectors leveraging these unique GAN fingerprints is important, as the misuse of such fakes (often referred to as deepfakes) has considerable potential harm to society, especially regarding misinformation and fraud \cite{ricker2024detectiondiffusionmodeldeepfakes}.

Early-stage GAN fake detection methods generally focused on the physical (or spatial) domain and directly on raw pixel intensities using convolutional neural networks (CNN), meaning that researchers developed a CNN for fakes and learned what suspects were and how the generative process left subtle cues. Many approaches to detection have reported high in-distribution accuracy, namely, perfect detection could be achieved when the detector was evaluated on the same type of fakes it was trained on. However, these spatial models could pick up on superficial cues that do not generalize well, and different types of GANs may have spatial artifacts that can differ \cite{wang2023generalgangeneratedimagedetection, wang2020cnn}. These challenges have led researchers to look for solutions derived from domains other than spatial.

Novel literature proposes that examining images across distinct domains may reveal robust artifacts that were less apparent in the spatial domain analysis. Prior studies demonstrated that applying image transforms such as 2D Fourier, Wavelet, and Median filters amplified artifacts that were missing in the spatial domain \cite{computers13120341, erukude2025identifying}. Moreover, training neural networks with these transformed images resulted in classifiers that are robust and generalizable. GANs do not reproduce the spectral properties of natural images in their entirety, particularly in the higher frequencies, because of the GAN up-sampling. These patterns appear in the Fourier spectrum as up-sampling artifacts (e.g., copying of frequency components)  \cite{info15110711, durall2020watch}. 

A detection framework that classified images based on their frequency spectrum instead of pixel representation was introduced \cite{zhang2019detecting}. A 2D DFT was applied to each image, followed by a log-transformation to the resulting magnitude spectrum, and then the three-channel frequency representation into a ResNet-based classifier. This method achieved cutting-edge performance in detecting GAN artifacts from fake images across multiple GAN models \cite{mahara2025methodstrendsdetectinggenerated}. Studies have shown that it can be approached using relatively straightforward spectral features (over 99\% accuracy), which model the decline of high-frequency Fourier coefficients \cite{mahara2025methodstrendsdetectinggenerated}. Other researchers have also looked to either wavelet transforms (DWT) or discrete cosine transforms (DCT) to detect frequency anomalies and/or match spatial with frequency streams in a unified detection network.

These studies collectively show that frequency-based trails (``GAN fingerprints") are a viable detection cue. It has even been demonstrated that detectors based on these cues can perform either on par or outperform deep spatial models; e.g., a shallow SVM classifier on Fourier features performed similarly or outperformed a deep CNN classifier that operates on pixel images.

Therefore, there is a pressing need for robust and generalizable detection pipelines that can reliably distinguish GAN-generated images from authentic ones. This study addresses this need by investigating the frequency domain as a source of subtle but consistent `fingerprints' left by GAN upsampling operations. By leveraging the 2D Discrete Fourier Transform and a ResNet50 classifier, our approach aims to reveal and exploit these hidden periodic artifacts to enhance detection accuracy and reliability.

\section{Data and CNN Architecture}

The dataset contains two primary semantic classes: human-face images and cat images. Within those classes, an equal number of real and GAN-generated (fake) images were acquired to facilitate robust training and evaluation of image authenticity detection models.

\subsection{Real Images Collection}

For Real Face Images, 2,500 images were randomly selected from the Flickr-Faces-HQ (FFHQ) \cite{ffhq_dataset} human faces. Altogether, FFHQ has 70,000 high-definition human face images with high definition, substantial demographic variability. FFHQ has great variability in age, gender, ethnicity, facial expression, and backgrounds, making it a suitable dataset that mimics the reality of diverse human faces. All images were resized to 256×256 for consistency.

\begin{figure}[ht]
    \centering
    \includegraphics[width=3in]{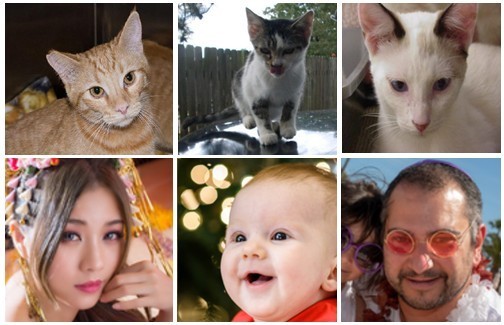}
    \caption{A sample of Real images from the dataset, including human faces (from FFHQ) and cat images (from the Cats vs. Dogs dataset). These images are the true class for the binary classification task. They also provide variety in terms of age, appearance, and background.}
    \label{fig_real_images}
\end{figure}

For Real Cat Images, 2,500 images were randomly selected from the Cats vs Dogs dataset on Kaggle \cite{cats_vs_dogs_kaggle}, which had been filtered to class cat images. This dataset provided a sufficient assortment of natural, real-world cat images to complete the second class of this dataset. Figure \ref{fig_real_images} presents a sample of images taken from the dataset that shows real cats and human faces.

\subsection{GAN-generated Images Collection}

Fake Face Images were produced by the StyleGAN2 architecture 
\cite{karras2020analyzingimprovingimagequality}. 2,500 images randomly selected from a publicly available collection of StyleGAN2-generated faces \cite{stylegan2_github}. The faces are very similar to real humans, and distinguishable artifacts are present from the GAN generation process.

\begin{figure}[ht]
    \centering
    \includegraphics[width=3in]{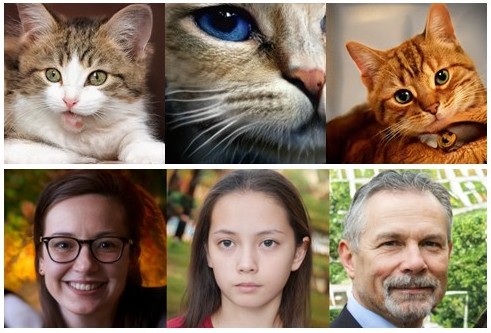}
    \caption{Sample GAN-generated (fake) images made by StyleGAN2, including synthetic human faces and images of cats. While they are visually realistic, these images contain artifacts (GAN fingerprints), which can be leveraged by deep learning models in the frequency domain for detection.}
    \label{fig_gan_images}
\end{figure}

A random sample of 2,500 Fake Cat Images was acquired similarly from an existing StyleGAN2-generated synthetic cat image collection to reflect both the distribution and resolutions of real cat images \cite{stylegan2_github}. Figure \ref{fig_gan_images} demonstrates a sample of GAN-generated images showing fake cats and human faces.

No pre- and post-processing processes were used on the images produced by GANs (except resizing when necessary). StyleGAN2 can produce a highly realistic image, but its use of convolutional up-sampling creates fine frequency artifacts that can be undetectable to the human eye but can be detected through spectral analysis in the frequency domain.

The dataset includes both human faces and animals, providing a meaningful yardstick for the potential generalizability of real vs. fake image detectors across various object classes and domains.

\subsection{Dataset Composition}

The overall dataset contained an equal distribution of real and fake images. A total of 10,000 images were collected, divided in a 70:15:15 ratio among the training (7,000), validation (1,500), and testing (1,500) sets, as shown in the Table \ref{tab:dataset_split_detailed}. The test set of 1,500 images was held out entirely during training. The dataset even contained an equal number of cat images and human faces. This 50/50 class balance is maintained to avoid biasing the classifier toward either class. Moreover, data augmentation was performed to enhance training diversity and model robustness. Images are rescaled to normalize pixel values. Random rotations (up to 15°), along with width and height shifts of up to 10\%, and zoom variations of up to 10\%. simulate real-world variations in object positioning and size. Horizontal flipping helps the model learn from mirrored images, while the ``reflect" fill mode preserves edge continuity during transformations.

\begin{table}[ht]
\centering
\caption{Detailed Dataset Composition and Split}
\label{tab:dataset_split_detailed}
\begin{tabular}{|l|c|c|c|c|}
\hline
\textbf{Subset}      & \textbf{Real Images} & \textbf{StyleGAN2 Images} & \textbf{Total Images} \\ \hline
Training Set         & 3,500                & 3,500                     & 7,000                 \\ 
Validation Set       & 750                  & 750                       & 1,500                 \\ 
Test Set            & 750                  & 750                       & 1,500                 \\ \hline
\textbf{Overall Total} & 5,000       &      5,000            &          10,000       \\ \hline
\end{tabular}
\end{table}

\subsection{CNN Architecture}

ResNet50 was utilized as the classifier, which is a 50-layer deep residual network that has demonstrated strong image classification capabilities across various computer vision applications \cite{he2015deepresiduallearningimage}. 

\begin{figure*}[ht]
    \centering
    \includegraphics[width=0.9\textwidth]{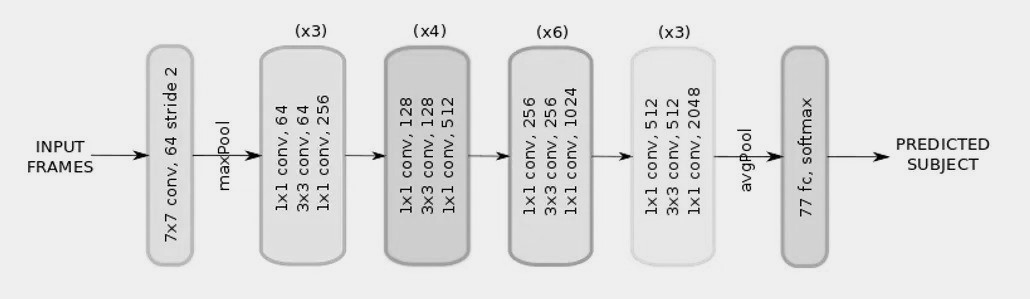}
    \caption{High-level architecture of the ResNet50 deep neural network used for both spatial-domain and frequency-domain classification tasks.}
    \label{fig_resnet}
\end{figure*}

Figure \ref{fig_resnet} depicts a high-level architecture of the network. ResNet50 was chosen as it is capable of learning subtle hidden patterns and artifacts in the spatial and frequency domains. Notably, the architecture and the hyperparameters were kept consistent throughout all the experiments to ensure a fair comparison between the spatial and frequency domain models.

The architecture is composed of stacked residual blocks, which have shortcut (skip) connections that let gradients skip over several convolutional layers when back-propagating. This alleviated the vanishing gradient problem and made it feasible to train much deeper networks. ResNet50 strikes a very good middle ground between depth and computation resources, and extracts very efficient features without significantly taxing computation resources.

ResNet50 was used for this study explicitly because the architecture is particularly good at learning complex representations of fine-grained properties and representations that are hierarchical in structure using image data. Specifically, below the residual block, ResNet50 will extract more low-level properties, with the layers learning progressively more abstract representations, such as shapes, hierarchical structures, and spatial relationships. Such multi-scale feature learning is crucial for detecting subtle artifacts (periodic frequency distortions) introduced by GANs, which have the potential to be visible at many levels of detail in an image.

One advantage of ResNet50 is that it is fully modular and transferable. The structure of the model was the same across the spatial-domain and frequency-domain studies, which allowed for direct comparison, and also enabled us to evaluate the effect of the input representation independently, without confounding results with a change in architecture. In this study, the network was initialized with the pretrained weights available from ImageNet, used transfer learning, and then fine-tuned the model for a binary classification task (real vs. fake). This reduced the training time significantly and improved generalization, as the model weights had already learned general visual properties that are potentially representative of images.

Among some of the key training hyperparameters, a learning rate of $1 \times 10^{-4}$, Adam for the optimizer, and a batch size of 32 were used. To facilitate convergence and address overfitting issues, several standard callbacks were implemented: \texttt{ModelCheckpoint}, \texttt{EarlyStopping}, \texttt{ReduceLROnPlateau}. It is important to note that all of the same hyperparameters and training protocols are used between spatial and frequency studies to maintain the experimental rigor and provide results that could be defensibly compared.

\section{Proposed work}

\subsection{Fourier Transform}

Discrete Fourier Transform (DFT) is a mathematical operation that maps an image from the spatial domain, where the intensity of each pixel is mapped according to its location, into the frequency domain, where an image is expressed by the presence and strength of the associated spatial frequencies \cite{computers13120341, erukude2025identifying}. In the Fourier domain, each point corresponds to a different frequency, and after applying a typical shift operation (like \texttt{fftshift}), the low-frequency component is centered while the high-frequency components go outward.

\begin{figure}[ht]
    \centering
    \includegraphics[width=1.7in]{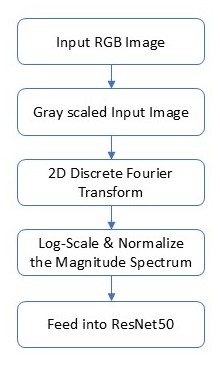}
    \caption{Flowchart illustrating the methodology for Fourier-based GAN fingerprint extraction and detection.}
    \label{fig_methodology}
\end{figure}

In image processing, low-frequency components typically contain information about smooth regions or generic outlines of objects, such as gradients or contours in the background, while the high-frequency components contain fine details, edges, or textures. Natural images commonly exhibit specific frequency distributions that inherently decrease in intensity with frequency (i.e., the energy fossilizes at higher frequencies), while GAN-generated images tend to fail in producing these exact statistical properties, particularly in the high-frequency bands, indicating violations of the natural correlation of frequencies. These failures result in evident periodic artifacts, spectral characteristics, or aliasing effects that can be used as discriminative characteristics of detection systems based on GAN models.

In this design, each image first undergoes 2D DFT, followed by the \texttt{fftshift} operation, which moves the zero-frequency component (DC) to the center, making the frequencies better interpretable. To assist in compressing the dynamic range of the image frequency magnitudes, which often span multiple orders of magnitude, the values are logarithmically transformed. The log-scaled values made mid- and high-frequency patterns more clearly visible, which would otherwise be dominated by the low-frequency energy. Finally, the log-transformed values are normalized into a $[0,1]$ range to accommodate deep learning architectures designed for normalized image inputs as shown in Figure \ref{fig_methodology}.

The favorable outcome of applying this approach in the frequency domain is that the eventual transformed image of the three color channels still has the same spatial dimensions as the original (256$\times$256) and can utilize commonly designed convolutional neural networks (CNNs) to avoid any modification to the overall architecture. CNNs exist to find local and hierarchical patterns and are capable to learn frequency-based signatures, such as GAN upscaling artifacts, aliasing induced by the convolution operation, or distinct periodic structures created during the generation of an image. If both assertive spatial arrangement and the corresponding spectral arrangement of energy are exploited, this model is better suited to distinguish real images from GAN images based on originally predicted characteristics, thus amplifying the robustness and generalizability of the detection \cite{computers13120341, erukude2025identifying}.

\begin{figure}[ht]
    \centering
    \includegraphics[width=3in]{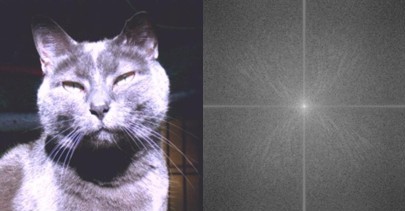}
    \caption{Example showing a real cat image in (left) the spatial domain (raw pixel intensity) and (right) the frequency domain (magnitude of the 2D DFT). The frequency domain can delineate energy in specific frequency bands, which may exhibit qualitatively different structures or randomness in comparison to the generated images from GANs, which is mostly relevant for detection.}
    \label{fig_fourier_cat}
\end{figure}

\subsection{Training Setup}

For both DFT-based and spatial-domain models, supervised learning is used to classify images as either ``Real" (class 0) or ``Fake" (i.e., StyleGAN2-generated, class 1). For the supervised learning approach, a binary cross-entropy loss function appropriate for binary classification was used; the models were optimized using the Adam optimizer with a learning rate of $1 \times 10^{-4}$. Adam is widely used and especially suited for a supervised learning approach because of its adaptive learning rate and momentum, which leads to faster convergence and better generalization than ordinary stochastic gradient descent.

The models were trained using a batch size of 32, which provided a balance between memory efficiency and gradient stability. Several data augmentation techniques, including random rotations (up to 15 degrees), width and height shifts (10\%), zoom (10\% in, 10\% out), and horizontal flips, were used to help the model be more robust. All augmentations are intended to help simulate the more commonly encountered real-world variation in image position, orientation, and scale for models to be invariant; this focused (and biased) training data helps narrow patterns they learn, but is not helpful for performance on real-world patterns.

To ensure that the model's training was stable and was not overfitting, the best practice in this regard was to include several commonly used callbacks: \texttt{ModelCheckpoint} (to save the best-performing model weights), \texttt{EarlyStopping} (to stop training when validation performance stops improving), and \texttt{ReduceLROnPlateau} (to reduce the learning rate after validation performance stops improving). The dataset was split into three datasets (training, validation, testing) using a 70:15:15 split (so that the test set was completely not observed during the training).

Most models were trained for a maximum of 100 epochs, although most training stopped during early stopping. All experiments were implemented using TensorFlow and most on an NVIDIA GPU, which provided fast training. Random seeds were set for all major libraries to provide reproducibility in this study. General model performance was assessed using accuracy, ROC AUC, and Average Precision (AP), which offered a comprehensive discussion of classifier quality across threshold-based and threshold-free measures. These findings align with prior studies that also reveal systematic discrepancies in the high-frequency Fourier modes of GAN-generated images, achieving over 99\% detection accuracy with spectral analysis \cite{dzanic2020fourier}.

\section{Result Analysis}

\begin{figure}[ht]
    \centering
    \includegraphics[width=3.7in]{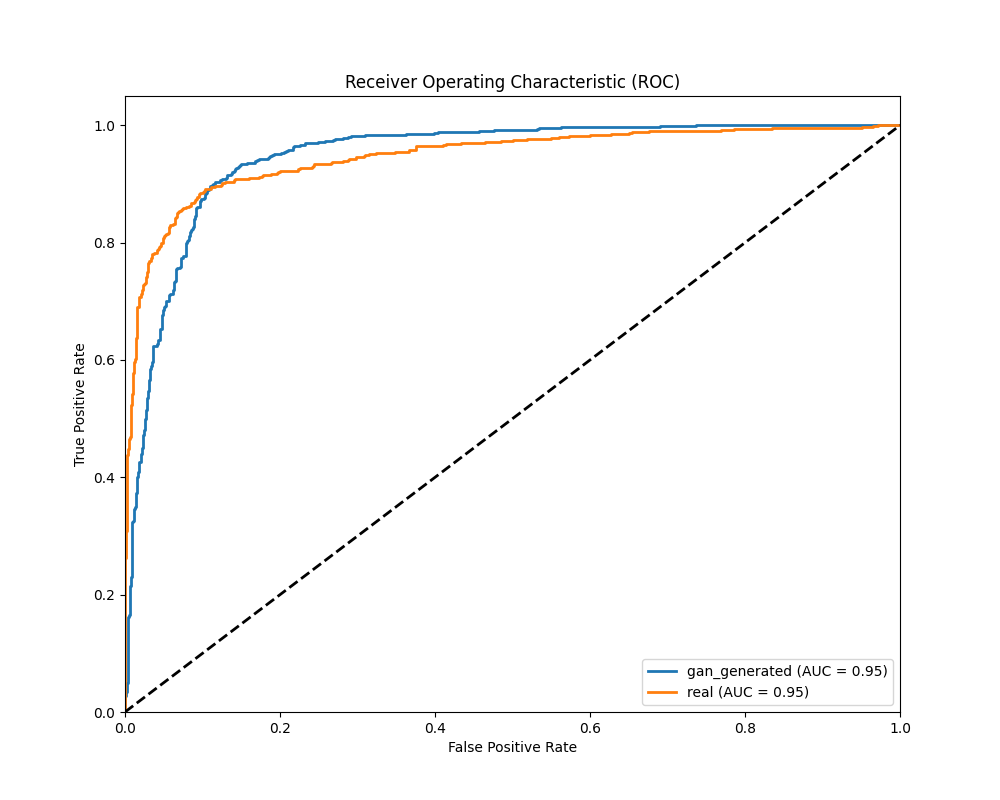}
    \caption{ROC curve for ResNet50 model trained and evaluated on the Fourier-transformed images.}
    \label{fig_roc_curve}
\end{figure}

After training, the spatial domain and frequency domain ResNet50 models were assessed on the 1,500-image test set. The DFT-based model significantly outperformed the spatial model on all metrics. The DFT model accuracy at 92.82\%, meaning it correctly identified real vs. fake images in the test set 92.82\% of the time, while the spatial model averaged 81.5\%, indicating a higher error rate when relying on raw pixels. The AUC improvement was comparable, for the frequency model 0.95 vs the spatial model 0.85. As depicted in the Figure \ref{fig_roc_curve}, an AUC of 0.95 means near separation of classes, vs 0.85, superior to random, leaving an overlap between the classes.  The average precision (AP), which is the harmonic mean of precision/recall trade-offs, was again the best approximation, with the DFT indexed at 0.95, and the spatial at 0.85, again implying a model with better confidence ranking on real vs fake images. Practically, using the standard decision threshold of 0.5, at DFT applied, there should be far fewer false positive and false negative cases than at the raw spatial pixels. The final cross-entropy loss on the test set was lower for the DFT model ($\approx$ 0.20 vs $\approx$ 0.33 for spatial), implying more conclusive predicted values should be expected at the frequency representation. The specific model metrics are summarized in Table \ref{tab:performance}.

\begin{figure}[ht]
    \centering
    \includegraphics[width=3.5in]{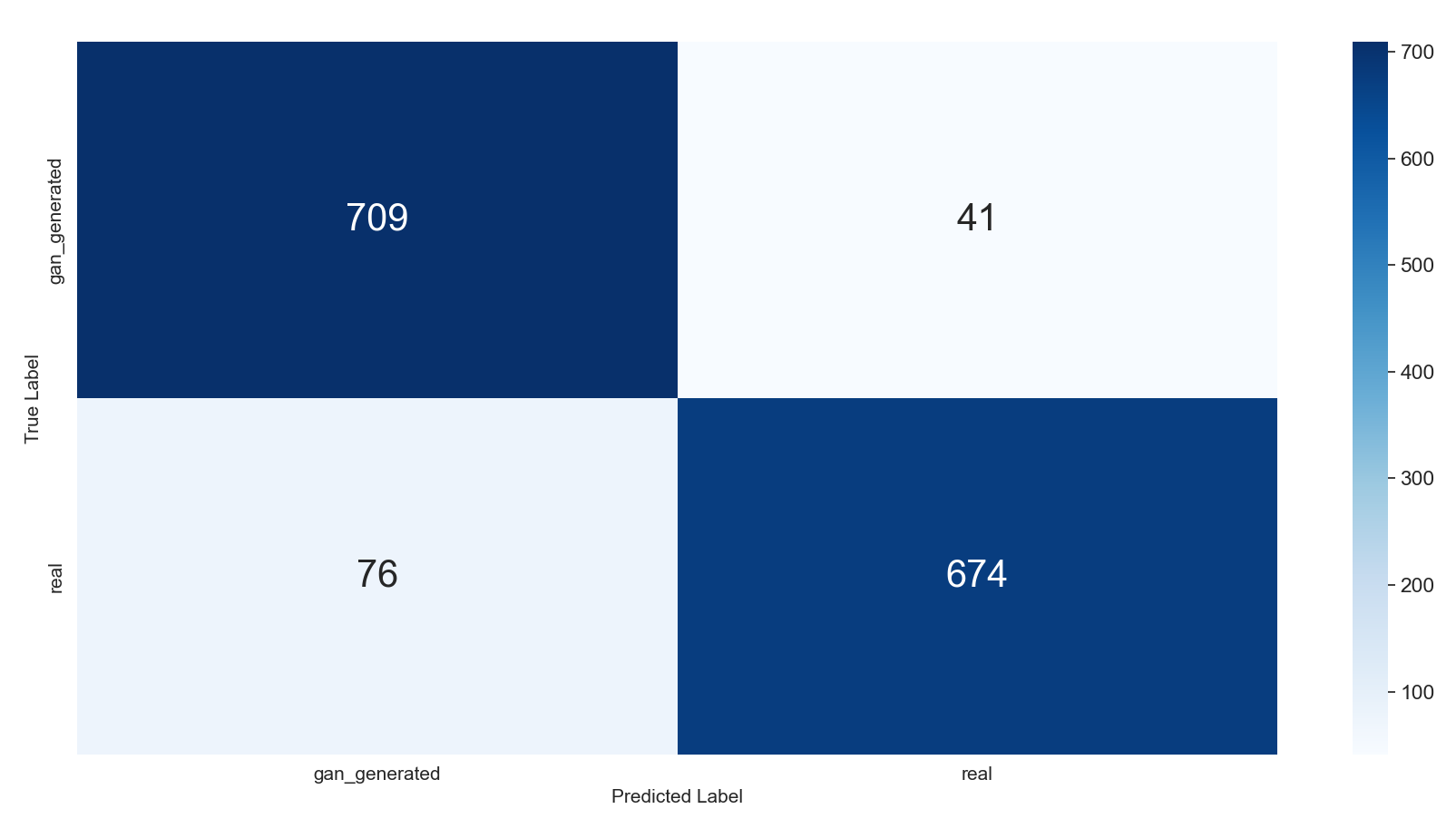}
    \caption{Confusion matrix showing the performance of DFT-RenNet50, with high true positive and true negative rates.}
    \label{fig_cm1}
\end{figure}

\begin{figure}[ht]
    \centering
    \includegraphics[width=3.5in]{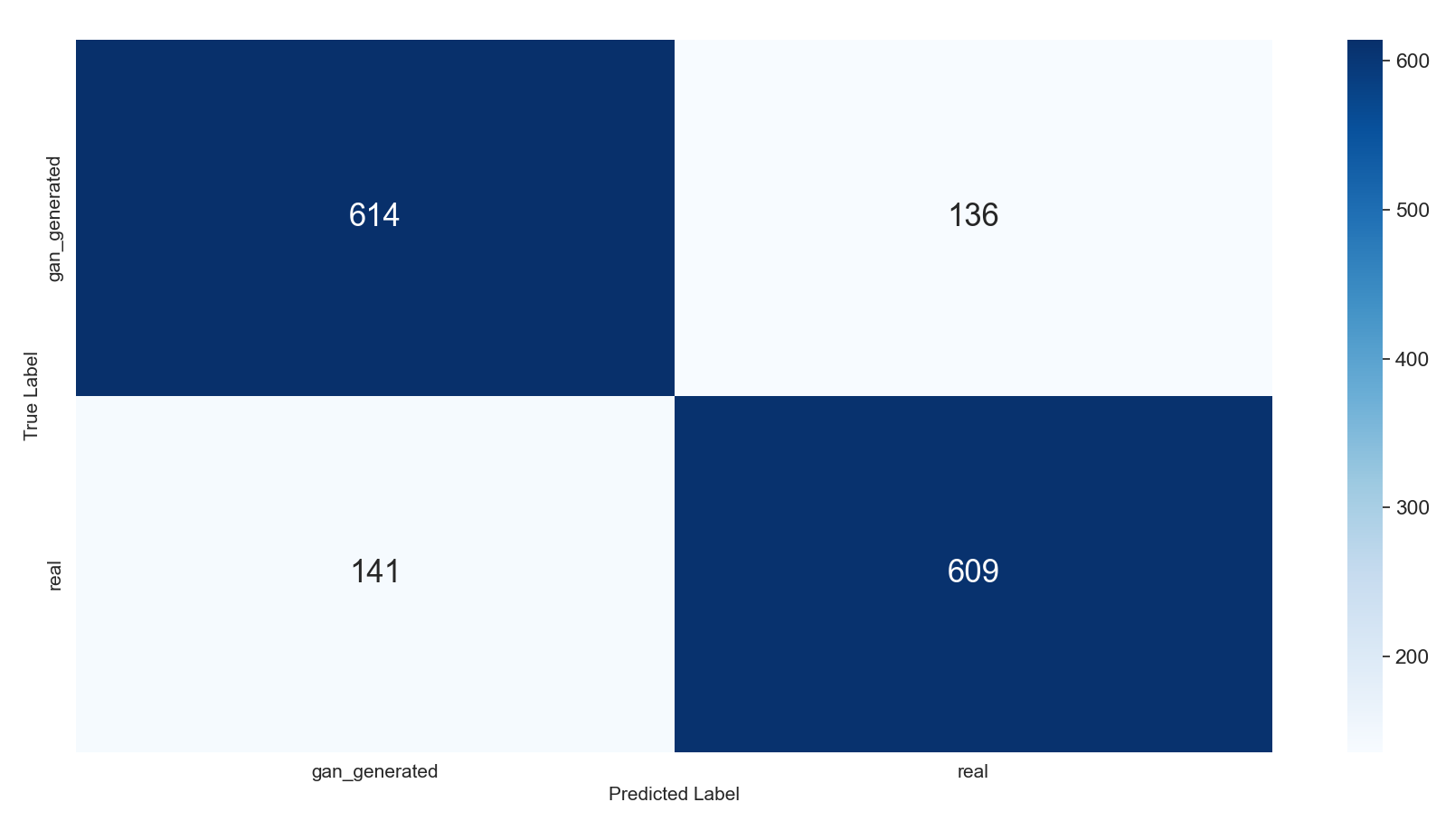}
    \caption{Confusion matrix showing the performance of Spatial-RenNet50, with relatively low true positive and true negative rates.}
    \label{fig_cm2}
\end{figure}

\begin{table}
\centering
\caption{Performance comparison on the test set of 1,500 images}
\label{tab:performance}
\begin{tabular}{|c|c|c|c|c|}
\hline
\textbf{Model} & \textbf{Accuracy (\%)} & \textbf{F1 Score} & \textbf{AUC} & \textbf{AP} \\
\hline
DFT-ResNet50          & 92.82   & 0.917      & 0.95          & 0.95 \\
Spatial-ResNet50      & 81.5    & 0.802      & 0.85          & 0.85          \\
\hline
\end{tabular}
\end{table}

Table \ref{tab:performance} also shows that the frequency-domain classifier had roughly an 11\% increase in accuracy over the spatial classifier. This difference is significant, especially considering the two models were trained on the same architecture, which tells us the frequency-domain input is providing a more useful signal for GAN detection. Figures \ref{fig_cm1}, \ref{fig_cm2} illustrate the confusion matrices of ResNet50 models trained on Fourier and Spatial data. The DFT-based model achieves higher true positive and true negative rates with fewer misclassifications. On the other hand, the spatial model has higher false negatives (failing to detect fakes) and false positives (mistaking real images for fakes). This highlights that frequency-domain analysis provides a stronger and more reliable signal for distinguishing GAN fingerprints compared to raw spatial features.

\section{Conclusion}

This paper presents a Fourier-based GAN-image detection approach and demonstrates its efficacy in distinguishing StyleGAN-generated images from real images. By transforming images into the frequency domain with 2D DFT, hidden periodic artifacts (``GAN fingerprints”) were exposed that a deep ResNet50 classifier could learn to recognize with high confidence. The frequency-domain model achieved 92.82\% accuracy and 0.95 AUC, substantially outperforming an equivalent ResNet50 operating on raw spatial images. These results confirm that frequency analysis provides a discriminative representation for GAN forensics, aligning with and reinforcing prior findings in the literature \cite{info15110711, zhang2019detecting}. The success of this experiment proves that even the most realistic GAN-generated images contain subtle hidden artifacts that differ from the real images in the Fourier domain. The findings from this paper have two major implications. The first, and more practical, is that frequency domain features may prove to be an attractive avenue to enhancing deepfake detection, especially for images. This could easily transfer to enhancing detection for imaging use cases such as social media image verification and counterfeit detection of digital content. The second, and much broader, implication from this paper is that it is important to keep studying the fingerprints of generative models. When distinct marks inserted by GANs or the signature modes of other generative models can be identified, it becomes possible to create better and potentially more generalized detectors.

Building on this work, future directions for advancing GAN image detection include testing cross-model generalization, expanding to multiple frequency representations like DWT and DCT, designing hybrid models that combine spatial and frequency domains, and improving robustness against common post-processing and adversarial attacks. Together, these efforts aim to keep forensic detection methods adaptable and effective as generative models evolve.

\section*{ACKNOWLEDGMENT}

The authors thank the contributors of the public datasets used in this study. The complete Python code has been made available at: \url{https://github.com/SaiTeja-Erukude/gan-fingerprint-detection-dft}.

\bibliographystyle{IEEEtran}
\bibliography{main}

\end{document}